\documentclass{article}
\usepackage{spconf,amsmath,graphicx}
\usepackage{bm}
\usepackage{multirow}
\usepackage{amsfonts}
\usepackage{color}

\def\L{{\cal L}}

\def \L{\mathcal{L}} 
\def \R{\mathbb{R}} 

\title{Self-Distilled Dynamic Fusion Network for Language-based Fashion Retrieval}
\name{$^{1}$Yiming Wu
        \qquad $^{1}$Hangfei Li
        \qquad $^{2}$Fangfang Wang
        \qquad $^{1}$Yilong Zhang
        \qquad $^{1}$Ronghua Liang}
  
  \address{$^{1}$ School of Computer Science and Technology, Zhejiang University of Technology, Zhejiang, China \\
      $^{2}$ Zhejiang Laboratory, Zhejiang, China}
\begin{document}
%
\maketitle

\begin{abstract}
    In the domain of language-based fashion image retrieval, pinpointing the desired fashion item using both a reference image and its accompanying textual description is an intriguing challenge. Existing approaches lean heavily on static fusion techniques, intertwining image and text. Despite their commendable advancements, these approaches are still limited by a deficiency in flexibility. In response, we propose a Self-distilled Dynamic Fusion Network to compose the multi-granularity features dynamically by considering the consistency of routing path and modality-specific information simultaneously. Two new modules are included in our proposed method: (1) Dynamic Fusion Network with Modality Specific Routers. The dynamic network enables a flexible determination of the routing for each reference image and modification text, taking into account their distinct semantics and distributions. (2) Self Path Distillation Loss. A stable path decision for queries benefits the optimization of feature extraction as well as routing, and we approach this by progressively refine the path decision with previous path information. Extensive experiments demonstrate the effectiveness of our proposed model compared to existing methods.
\end{abstract}


\section{Introduction}
\begin{figure}[t]
    \centering
    \includegraphics[width=0.85\linewidth]{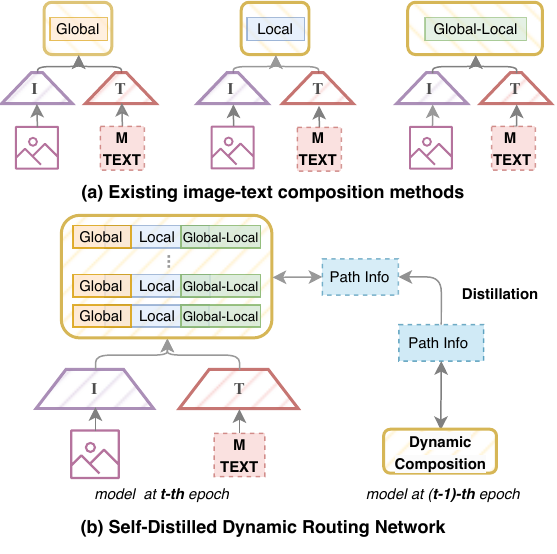}
    \caption{Difference between our proposed method and the existing composition methods.}
    \label{fig:comparison}
\end{figure}

Language-based fashion image retrieval is the task of targeting the specific image given the reference image and the accompanying modification text~\cite{Park_2019_CVPR_Workshops,Goenka_2022_fvlp_CVPR,Vo2018ComposingTA}. The bi-modality of the query allows more expressive and precise description of the target images, as some features are easier to be described with language, and others are more suitable for image. Thus, numerous methods~\cite{Noh_Araujo_Sim_Weyand_Han_2017} have been introduced to compose complementary queries and have made tremendous progress over the past few years. However, there exist some critical limitations. First, static feature composition modules hinder the generalization of the model in various queries, encompassing specific data distributions and query formats. Second, the current individual composition modules are not able to fully capture the intricate modalities at different granularity levels when merging image and text.

Motivated by the above observations, we introduce the Self-distilled Dynamic Fusion Network (SDFN), which dynamically composes multi-modality queries. As illustrated in Figure~\ref{fig:comparison}, we compare our proposed method with existing composition methods. Specifically, our proposed SDFN handles the feature composition problem by incorporating multiple operation modules, each of which is tailored to different specific information within queries. These fusion modules effectively fuse image and text representations of varying granularity, thereby facilitating comprehensive and distinctive modality fusion for each image-text pair. Meanwhile, the self-distilled dynamic fusion method enables dynamic feature fusion through the aforementioned operation modules, which simultaneously considers the consistency of the routing path and the modality-specific information. Due to the disparate modalities of the reference image and the modification text, it is imperative to employ routers of different architectures for the image and text to make path decisions separately. In light of this, we propose Modality Specific Routers (MSR) consisting of an image router and a text router. By amalgamating the routing probabilities generated by both routers, we can achieve more precise and accurate routing decisions.

Our main contributions are three-fold: 1) We propose a Self-distilled Dynamic Fusion Network (SDFN) which employs several operation modules to model a flexible and comprehensive modality interaction. By integrating Modality Specific Routers into each operation module, our SDFN can make path decisions based on the characteristics of both image and text modalities. 2) To stabilize path decisions and facilitate the optimization of feature extraction and routing processes, we introduce the self-path distillation (SPD), which utilizes historical path information to enhance the path the routing decision-making capabilities of routers. 3) We perform rigorous evaluations on three widely used CIR benchmarks, with the experimental results showing the superiority of our approach.

\section{Methodology}\label{sec:method}

\begin{figure}[t]
  \begin{center}
  \includegraphics[width=0.9\linewidth]{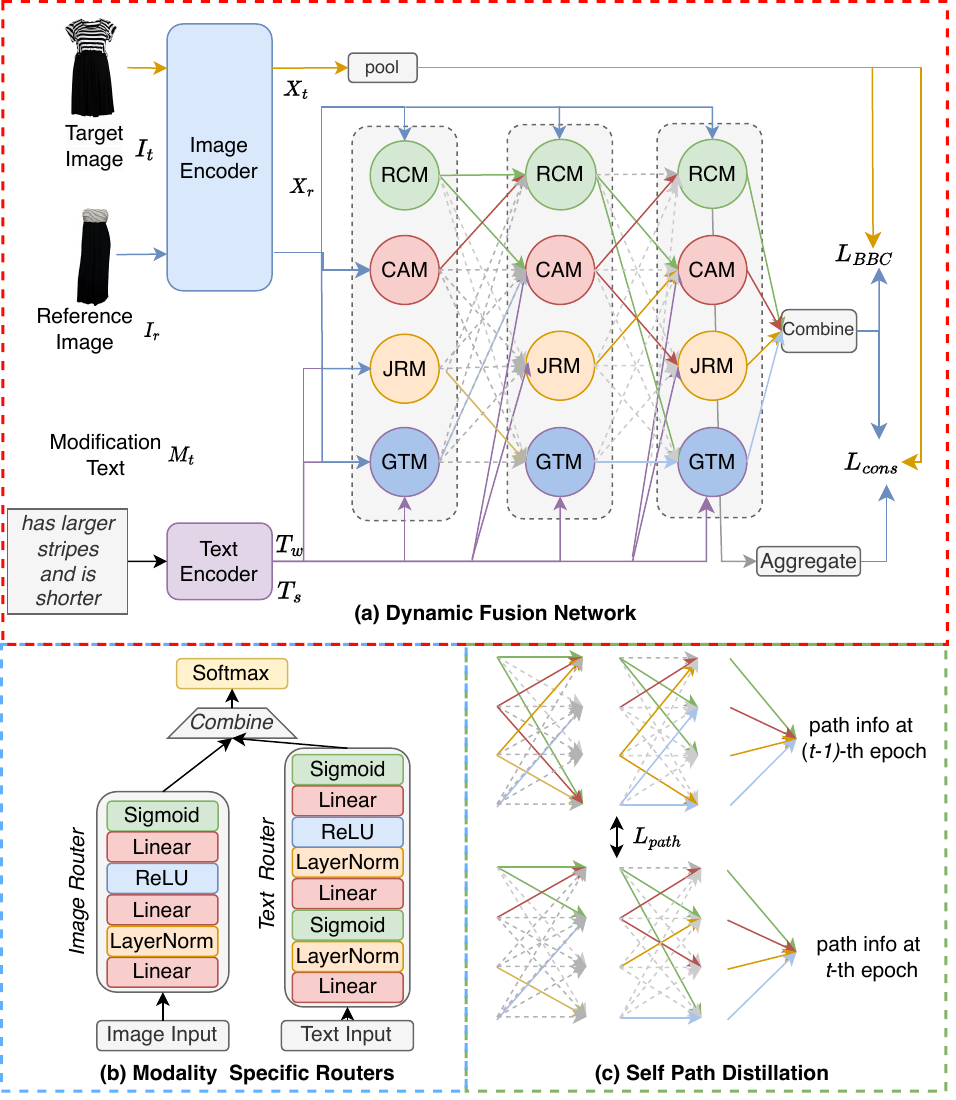}
  \end{center}
  \caption{Overall architecture of our proposed Self-distilled Dynamic Fusion Network. \textbf{(a)} Dynamic Fusion Network. \textbf{(b)} Modality Specific Routers. \textbf{(c)} Self-Path Distillation.}
  \label{fig:SDFN}
\end{figure}

Given a composed image retrieval dataset consisting of $N$ triplets $\mathcal{D}=\{(I_r^i,M_t^i,I_t^i)\}_{i=1}^N$, where $I_r^i$, $M_t^i$, and $I_t^i$ are the reference image, the modification text, and the target image, respectively. Our main objective is to learn a joint embedding space in which the distance between the image-text joint representation and the corresponding target image representation should be as close as possible.

\subsection{Image and Text Representation}
\noindent\textbf{Image Representation}. As illustrated in Figure~\ref{fig:SDFN} (a), image encoder $F_{img}$ is shared for reference image $I_r$ and the corresponding target image $I_t$, image features are extracted with feature extractor $X_r = F_{img}(I_r), X_t = F_{img}(I_t)$, where $X_r,X_t \in \R^{H\times W\times 2D}$ are the last feature map output from ResNet, and we apply $1\times 1$ convolution to reduce the dimension of the image feature. For the reference image feature, it is reshaped as $X_r\in\R^{K \times D}, K = H \times W$. Regarding the target image feature, we obtain the final target image feature $X_t\in\R^D$ with an average pooling operation.

\noindent\textbf{Text Representation}. To encode the modification text, we employ an LSTM as the text encoder. We first tokenize the text into a token sequence and then feed these tokens into LSTM and obtain the word-level representation $T_w = F_{txt}(M_t)$, where $M_t$ denotes the modification text, $T_w\in \R^{L\times D}$ and $L$ is the number of words in the caption. Then, we apply max pooling on the word-level representation to obtain the sentence-level representation $T_s\in\R^D$.

\subsection{Operation Modules}~\label{sec:operation modules}
The dynamic fusion network enables multi-granularity modality interaction through comprehensive and diverse operation modules. We employ four distinct image-text fusion modules based on several modality interaction strategies. Detailed descriptions are as follows~\footnote{For simplicity, we reuse some notations among different operation modules, the parameters across these modules are not shared.}.

\noindent\textbf{Joint Reasoning Module (JRM).} Within this module, we fuse the reference image feature $X_r$ and the coarse-grained sentence-level representation $T_s$. Specifically, $X_r$ and $T_s$ are concatenated and fed to the multi-head attention module for further comprehensive interaction. The process is formally represented as follows:
\begin{equation}
\begin{aligned}
    X_{cat} &= W_{cat} [X_r,T_s],\\
    X_{att} &= MHA(W_q X_{cat}, W_k X_{cat}, W_v X_{cat}),\\
    O_{jrm} &= FFN(X_{att}) + X_{att},
\end{aligned}
\end{equation}
where $[\cdot, \cdot]$ denotes the concatenate operation, and $W_{cat} \in \R^{D \times 2D}$ linearly transforms the concatenated features, and $FFN(\cdot)$ is the feedforward network.

\noindent\textbf{Cross Attention Module (CAM).}
We propose the Cross Attention Module to capture the fine-grained inter-modality relationship and perform the modality interaction between text descriptions and image regions. We fuse reference image feature $X_r$ and word-level representation $T_w$ as follows:
\begin{equation}
O_{cam} = LN(MHA(W_q X_r, W_k T_w, W_v T_w)),
\end{equation}
where $W_q, W_k, W_v \in \R^{D \times D}$ are parameters of linear layer, $LN(\cdot)$ and $MHA(\cdot)$ refers to layer normalization and multi-head attention, respectively.

\noindent\textbf{Global Transformation Module (GTM).}~\label{sec:gtm}
Previous modules perform interaction between image and text through an attention operation. In this module, we propose to project image representation with text features. We first project sentence-level text feature to generate a scaling vector and a shifting vector $\alpha = W_{\alpha} T_s, \beta = W_{\beta} T_s$, where $W_{\alpha}, W_{\beta}\in\R^{D\times D}$ are the learnable transformation matrices. Then we utilize these two vectors to perform global transformation for image features $O_{gtm} = LN(\alpha X_r + \beta)$.

\noindent\textbf{Residual Connection Module (RCM).}~\label{sec:rcm}

The introduction of the residual module is motivated by two primary factors. First, we observe that some reference images and the corresponding target images share high visual similarity, it is imperative to retain the semantics of the reference image to the greatest extent. Second, the residual connection shows benefits in mitigating feature collapse in stacked layers. For these reasons, we propose the residual connection module, which is simply formulated as $O_{rcm} = LN(X_r)$.

\subsection{Modality Specific Routers}~\label{sec:modality specific routers}\\
 
As depicted in Figure~\ref{fig:SDFN} (b), we propose two separate routers for image and text.

In the dynamic fusion network, the above operation modules receive the reference image feature and the text feature, and transmit the fused features coupled with routing probabilities from routers to the next layer as the reference image feature, this routing process is formulated as follows,
\begin{equation}
    X_i^l  = 
    \begin{cases}
        X_r  \quad\quad\quad\quad\quad\quad\quad   l=0,\\
        \sum_{j\in S} O_j^{l-1} R_{j\rightarrow i}^{l-1} \quad l>0,
    \end{cases}
\end{equation}
where $i, j \in \{cam, jrm, gtm, rcm\}$ is the name of operation module, $X_i^l$ denotes the the input of the $i$ module in the $l$-th layer, and $R_{j\rightarrow i}^{l-1}$ indicates the routing probability from the $j$ module in the $(l-1)$-th layer to the $i$ module in $l$-th layer, $O_j^{l-1}$ is the output of the $j$ module in the $(l-1)$-th layer.

\subsection{Objective Function}\label{sec:objective function}
For training, we employ self path distillation loss, batch-base classification loss~\cite{Vo2018ComposingTA} and consistency loss.

\noindent\textbf{Self Path Distillation Loss}. 
We propose self path distillation to enhance the stability of decision making by refining the softened path predictions. As shown in Figure~\ref{fig:SDFN} (c), SPD employs the model from preceding epoch as the teacher and transfers path information to the model in the current epoch. Formally, the loss function is written as:
\begin{equation}
    \L_{path} = \frac{1}{B}\sum_{i=1}^{B}\tau_{path}^2\cdot D_{KL}(p_s||p_t),
\end{equation}
where $p_s,p_t$ are the softened predictions from the student and the teacher, respectively. And $\tau_{path}$ is the temperature parameter, $D_{KL}$ denotes the Kullback-Leibler (KL) divergence.

\noindent\textbf{Consistency Loss.}

To maintain the consistency and robustness of representations before and after multi-layer fusion, we introduce the consistency loss $L_{cons}$ as follows,
\begin{equation}
    M_q = f_qf_q^\top,\quad
    M_t = f_tf_t^\top, \quad
    M_{in} = f_{in}f_{in}^\top,
\end{equation}
\begin{equation}
   \L_{cons} = ||M_q,M_t||_2 + ||M_{in},M_t||_2,
\end{equation}
where $f_q\in\R^{B\times D}$ is the output features for the last fusion layer, $f_t\in\R^{B\times D}$ is the target image features, and $f_{in}\in\R^{B\times D}$ denotes the summation of the pooled input features for the last fusion layer. Our final optimization objective is:
\begin{equation}
    \L_{total} = \L_{BBC} + \L_{cons} + \lambda \L_{path},
\label{eq:loss}
\end{equation}
where $L_{BBC}$ is the batch-base classification loss.

\section{Experiments}\label{sec:experiments}
\subsection{Experiment setting}\label{sec:experiment setting}
\noindent\textbf{Dataset and Protocal.} We assess our proposed method on three benchmarks: FashionIQ~\cite{Wu_iq_2021}, Shoes~\cite{Berg_shoes_2010}, and Fashion200k~\cite{Vo2018ComposingTA}, adhering to the standard evaluation protocols adopted in ~\cite{Vo2018ComposingTA,Lee_cosmo_2021,Delmas2022ARTEMISAR,Wen_clvcnet_2021}. For FashionIQ, we compute the R10/50 across categories and present an average from six subcategories. For Shoes and Fashion200k, we provide the R1/10/50 metrics and their average for comparison.

\noindent\textbf{Implementation Details.} We employ ResNet-50 (pretrained) and LSTM for image and text encoding, respectively. The final image representation is derived from ResNet-50's last feature map, with its dimensionality reduced from 2048 to 1024. For text, GloVe tokenizes words, and an LSTM with 1024 hidden units finalizes the representation. For training optimization, we utilize Adam with a weight decay of $10^{-6}$ and a learning rate of $10^{-4}$. Training specifics for each dataset are as follows:
FashionIQ: 60 epochs, batch size of 32, learning rate decays by a factor of 10 at epoch 50, and $\lambda=1$. Shoes: 30 epochs, batch size of 16, learning rate decays at epoch 15, and $\lambda=0.6$. Fashion200K: 50 epochs, batch size of 64, learning rate decays at epoch 30, and $\lambda=0.6$.
Our dynamic fusion network consists of three stacked operation module layers.

\begin{table}[t]
\resizebox{0.48\textwidth}{!}{
\begin{tabular}{lcccccccc}
\hline
\multirow{2}{*}{Method}                     & \multicolumn{4}{c}{Shoes}        & \multicolumn{4}{c}{Fashion200K}  \\ \cline{2-9} 
                                            & R1     & R10   & R50   & Avg     & R1   & R10  & R50  & Avg      \\ \hline
FiLM~\cite{Perez_film_2018}                 & 10.19  & 38.39 & 68.30 & 38.96   & 12.9 & 39.5 & 61.9 & 38.1         \\
TIRG~\cite{Vo2018ComposingTA}               & 12.60  & 45.45 & 69.39 & 42.48   & 14.1 & 42.5 & 63.8 & 40.1         \\
VAL~\cite{Chen_val_2020}                    & 16.49  & 49.10 & 71.45 & 42.95   & 21.2 & 49.0 & 68.8 & 46.3         \\
CoSMo~\cite{Lee_cosmo_2021}                 & 16.72  & 48.36 & 75.64 & 46.91   & 23.3 & 50.4 & 69.3 & 47.6         \\
MAAF~\cite{Dodds_maaf_2020}                 & 16.45  & 49.95 & 76.36 & 47.58   & -    & -    & -    & -            \\
DATIR~\cite{Gu2021datirSW}                  & 17.20  & 51.10 & 75.60 & 47.97   & -    & -    & -    & -            \\
FashionVLP~\cite{Goenka_2022_fvlp_CVPR}     & -      & 49.08 & 77.32 & -       & -    & 49.9 & 70.5 & -            \\
MCR~\cite{Zhang2021mcr}                     & 17.85  & 50.95 & 77.24 & 48.68   & -    & -    & -    & -            \\
SAC ~\textit{w/ BERT}~\cite{Jandial2020SAC} & 18.50  & 51.73 & 77.28 & 49.17   & -    & -    & -    & -            \\
ARTEMIS~\cite{Delmas2022ARTEMISAR}          & 18.72  & 53.11 & 79.31 & 50.38   & 21.5 & 51.1 & 70.5 & 47.7         \\
DCNet~\cite{Kim_dcnet_2022}                 & -      & 53.82 & 79.33 & -       & -    & 46.9 & 67.5 & -             \\
CLVC-Net*~\cite{Wen_clvcnet_2021}           & 17.64  & 54.39 & 79.47 & 50.50   & 22.6 & \textbf{53.0} & 72.2 & 49.2    \\
EER~\cite{Zhang2022eer}                     & 20.05  & 56.02 & 79.94 & 52.00   & -    & -    & -    & -            \\
AMC~\cite{Zhu_amc_Trans_Multimedia_Commun_Appl} & 19.19  & 56.89 & 79.27 & 52.05 & -    & -    & -    & -            \\
Css-Net$*$~\cite{Zhang_cssnet_Wang}        & 20.13  & 56.81 &\textbf{81.32} & 52.75  & 23.4 & 52.0  & 72.0 & 49.1  \\ \hline
\textbf{SDFN(Ours)}                         &\textbf{22.07} &\textbf{57.56} & 80.88 & \textbf{53.53} & \textbf{26.3} & 51.1          & \textbf{72.6} & \textbf{50.0}\\ \hline
\end{tabular}
}
\caption{Performance comparison on Shoes and Fashion200K. $*$ denotes model ensemble.}
\label{tab: comparison on shoes and fashion200k}
\end{table}

\begin{table}
\resizebox{0.48\textwidth}{!}{
\begin{tabular}{lcccccccccc}
\hline
\multicolumn{1}{c}{\multirow{2}{*}{Method}}        &\multicolumn{2}{c}{Dress} &\multicolumn{2}{c}{Shirt} &\multicolumn{2}{c}{Toptee} &\multicolumn{2}{c}{Avg} &\multicolumn{2}{c}{\multirow{2}{*}{Avg}} \\ \cline{2-9}
\multicolumn{1}{c}{}        & R10   & R50    & R10   & R50   & R10   & R50    & R10    & R50     &\multicolumn{1}{c}{} \\ \hline
MRN~\cite{Kim_mrn_2016}     & 12.32 & 32.18  & 15.88 & 34.33 & 18.11 & 36.33  & 15.44  & 34.28   &\multicolumn{2}{c}{24.86} \\
FiLM~\cite{Perez_film_2018} & 14.23  & 33.34 & 15.04 & 34.09 & 17.30 & 37.68  & 15.52  & 35.04   &\multicolumn{2}{c}{25.28}  \\
TIRG~\cite{Vo2018ComposingTA}& 14.87 & 34.66 & 18.26 & 37.89 & 19.08 & 39.62  & 17.04  & 37.39   &\multicolumn{2}{c}{27.40}   \\
VAL~\cite{Chen_val_2020}     & 22.53 & 44.00 & 22.38 & 44.15 & 27.53 & 51.68  & 24.15  & 46.61   &\multicolumn{2}{c}{35.40}  \\
CoSMo~\cite{Lee_cosmo_2021}  & 26.45 & 52.43 & 26.94 & 52.99 & 31.95 & 62.09  & 28.45  & 55.84   &\multicolumn{2}{c}{39.45}  \\
AMC~\cite{Zhu_amc_Trans_Multimedia_Commun_Appl} & 31.73  & 59.25 & 30.67  & 59.08  & 36.21 & 66.00 & 32.87 & 61.64 & \multicolumn{2}{c}{47.25} \\
FashionVLP~\cite{Goenka_2022_fvlp_CVPR}  & 32.42 & 60.29 & 31.89 & 58.44 & 38.51 & 68.79 & 34.27 & 62.51 & \multicolumn{2}{c}{48.39}    \\
LIMN~\cite{Wen_limn_Wen}     & 35.60 & 62.37 & 34.69 & 59.81 & 40.64 & 68.33 & 36.98   & 63.50   &\multicolumn{2}{c}{50.24}  \\
ComqueryFormer~\cite{Xu_comqueryformer_2023} & 33.86 & 61.08 & 35.57 & 62.19   & 42.07  & 69.30  & 37.17 & 64.19 & \multicolumn{2}{c}{50.68}  \\
Css-Net~\cite{Zhang_cssnet_Wang} & 33.65 & 63.16  & \textbf{35.96} & 61.69  & 42.65   & 70.70 & 37.42   & 65.27   & \multicolumn{2}{c}{51.34}                    \\ \hline
\textbf{SDFN(Ours)}  & \textbf{37.48} & \textbf{65.15} & 35.82 & \textbf{62.32} & \textbf{44.42} & \textbf{71.44}

& \textbf{39.24} & \textbf{66.30} & \multicolumn{2}{c}{\textbf{52.77}} \\ \hline \\
\end{tabular}
}
\caption{Performance comparison on FashionIQ dataset. The results of VAL Evaluation protocol. $*$ indicates model ensemble.}
\label{tab:comparison on iq}
\end{table}

\subsection{Main Results}
The main results are reported in Table~\ref{tab: comparison on shoes and fashion200k} and~\ref{tab:comparison on iq}. On Shoes dataset, our model achieves the remarkable performance and surpasses the Css-Net. While our proposed model is slightly behind Css-Net on the R50, it obtains improvements of +1.94\% R1, +1.05\% R10 and +0.73\% on average. On Fashion200K dataset, our model surpasses Css-Net* and CLVC-Net* which apply model ensemble to further boost the performance. Specifically, our model obtain an improvement of 3.7\% and 0.4\% on R1 and R50 compared with CVLC-Net* and achieve an improvement of 2.9\% and 0.6\% on R1 and R50 compared with Css-Net*. As for FashionIQ dataset, we report the results with VAL evaluation protocol. Our model outperforms existing models by a significant margin. Compared to Css-Net, our model achieves improvements of +1.82\% R10 and +1.03\% R50 on average.

\begin{table}[t]
\resizebox{0.48\textwidth}{!}{
\begin{tabular}{lcccccc}
\hline
\multicolumn{2}{l}{\multirow{2}{*}{Model}}          & \multicolumn{3}{c}{Shoes} & \multicolumn{2}{c}{FashionIQ (Avg)} \\ \cline{3-7} 
\multicolumn{2}{c}{}                                & R1     & R10   & R50   & R10             & R50            \\ \hline
\multicolumn{2}{l}{Baseline}                        & 16.04   & 50.63  & 77.98  & 27.42            & 54.65           \\
\multicolumn{2}{l}{Baseline w/SR}                  & 19.39   & 56.83  & 79.46  & 28.41            & 55.36           \\
\multicolumn{2}{l}{Baseline w/MSR}                 & 20.67   & 55.32  & 79.55  & 29.58            & 56.75           \\
\multicolumn{2}{l}{Baseline w/~\textit{$L_{cons}$}}             & 16.84   & 52.96  & 78.24  & 28.82            & 55.12           \\
\multicolumn{2}{l}{Baseline w/~\textit{$L_{cons}$} + SR}        & 20.22   & 56.48  & 79.83  & 29.34            & 56.01           \\
\multicolumn{2}{l}{Baseline w/~\textit{$L_{cons}$} + MSR}       & 21.84   & 56.91  & 80.60  & 29.58            & 56.75           \\
\multicolumn{2}{l}{Baseline w/~\textit{$L_{cons}$} + MSR + SPD (SDFN)} & ~\textbf{22.06}   & ~\textbf{57.65}  &~\textbf{80.88} & ~\textbf{31.30}            & ~\textbf{57.32}       \\ \hline  
  \multicolumn{6}{l}{\textcolor[rgb]{0.502,0.502,0.502}{impact of each operation module}}                                                                          \\ 
\hline
\multicolumn{2}{l}{SDFN w/o RSM} & 14.28   & 48.45  & 75.51 & 28.38            & 55.13      \\
\multicolumn{2}{l}{SDFN w/o JRM} & 20.10   & 55.43  &79.12    & 29.76       & 56.01     \\
\multicolumn{2}{l}{SDFN w/o GTM} & 20.24   & 56.03  & 79.35 & 29.08            & 55.65       \\ 
\multicolumn{2}{l}{SDFN w/o CAM} & 20.19   & 54.81  & 79.18 & 29.01            & 55.69      \\ 

\hline  
\end{tabular}
}
\caption{Evaluation of individual component on Shoes and FashionIQ datasets.}
\label{tab:ablation}
\end{table}

\subsection{Ablation Studies}\label{sec:ablation studies}
We present the results of ablation study in Table~\ref{tab:ablation}, in which $L_{cons}$ represents consistency loss, SR denotes modality-agnostic router for both image and text features, MSR is our proposed Modality Specific Routers, SPD means Self Path Distillation loss, 
RCM/CAM/GTM/JRM are operation modules in the dynamic fusion network.

\noindent \textbf{Effectiveness of Modality Specific Routers}. It can be observed that the models w/SR perform worse than those w/MSR, regardless of whether the consistency loss is added or not, which validates the effectiveness of our proposed MSR, providing evidence for the existence of a pronounced modality gap and significant impact on the retrieval accuracy.

\noindent \textbf{Impact of consistency loss}. By comparing Baseline and Basline w/ $L_{cons}$, we could find that the consistency loss leads to substantial performance improvements on the Shoes and FashionIQ datasets.

\noindent \textbf{Effectiveness of Self Path Distillation}. The path decision process is considered as a classification task aimed at identifying optimal routing combinations for individual input queries. SPD utilizes the path predictions generated in previous epochs to generate high-quality soft targets. From Table~\ref{tab:ablation}, we could see that SPD improves performance compared to the result in the penultimate row. 
 This demonstrates the effectiveness of knowledge acquisition through self-distillation from the model of the last epoch.

\noindent \textbf{Impact of operation modules}. We conduct a validation of the effectiveness of each operation module in our SDRN. The results, as presented in Table~\ref{tab:ablation}, indicate that the removal of the residual module has the most significant impact on the model performance. Specifically, it leads to a decrease of 7.78\%, 9.2\%, and 5.37\% in R1, R10, and R50, respectively, on the Shoes dataset. One possible reason is that the residual module provides more inter-layer interaction, thereby alleviating the issue of missing feature diversity due to multi-layer modality interactions. Removal of JRM and CAM results in similar performance degradation, while GTM exhibits the smallest impact compared to other modules.

\section{Conclusion}
In this work, we propose a self-distilled dynamic fusion network for CIR task. By seamlessly integrating modality specific routers with several operation module, our proposed SDFN constructs diverse routing graph for different queries. In order to maintain the stability and accuracy of routing process, we employ self path distillation to transfer the path knowledge from previous epoch to further improve model performance. We validate our model on three benchmarks and demonstrate the effectiveness of our proposed method.

\vfill\pagebreak
\label{sec:refs}
\bibliographystyle{IEEEbib}
\bibliography{strings}

\end{document}